\documentclass[letterpaper,10pt,journal,twoside]{IEEEtran}

\IEEEoverridecommandlockouts

\usepackage{cite}
\usepackage{amsmath,amssymb,amsfonts}
\usepackage{algorithmic}
\usepackage{graphicx}
\usepackage{textcomp}
\usepackage{xcolor}
\usepackage{xurl}
\usepackage{microtype}
\usepackage[normalem]{ulem}

\newif\ifrevision
\revisionfalse  

\ifrevision
  \newcommand{\rev}[1]{{\color{blue}#1}}
  \newcommand{\revdel}[1]{{\color{red}\sout{#1}}}
\else
  \newcommand{\rev}[1]{#1}
  \newcommand{\revdel}[1]{}
\fi

\begin{document}

\title{\resizebox{\textwidth}{!}{\shortstack{Ground Reaction Force-Aware Model Predictive Control\\[0.5ex]for Underactuated Upright Physical Assistance}}}

\author{Darren Biskup\textsuperscript{1}, Jeonghan Yu\textsuperscript{1,2}, Zibo Yu\textsuperscript{1}, and Sunil K. Agrawal\textsuperscript{1}%
\thanks{Manuscript received: May 31, 2026;
Revised: September 15, 2026;
Accepted: October 13, 2026.}%
\thanks{This paper was recommended for publication by Editor Pietro Valdastri upon evaluation of the Associate Editor and
Reviewers' comments. This work was supported in part by the
New York Spinal Cord Injury Research Board under Grant
DOH01-C37719GG-3450000 and in part by the Columbia-Dream Sports
Artificial Intelligence (AI) Innovation Center.
\textit{(Jeonghan Yu and Zibo Yu contributed equally to this work.)}
\textit{(Corresponding author: Sunil K. Agrawal.)}}%
\thanks{\textsuperscript{1}Darren Biskup, Zibo Yu, and Sunil K. Agrawal
are with the Department of Mechanical Engineering, Columbia University,
New York, NY 10027, USA. Jeonghan Yu was also with this department (e-mail: db3728@columbia.edu; zy2753@columbia.edu;
sa3077@columbia.edu).}%
\thanks{\textsuperscript{2}Jeonghan Yu is with the Department of Mechanical Engineering,
SUNY Korea, Incheon 21985, Republic of Korea (e-mail: jeonghan.yu@sunykorea.ac.kr).}%
\thanks{Sunil K. Agrawal is also with the Department of
Rehabilitation and Regenerative Medicine, Columbia University,
New York, NY 10027, USA.}%
\thanks{\textcopyright{} 2026 IEEE.  Personal use of this material is permitted.  Permission from IEEE must be obtained for all other uses, in any current or future media, including reprinting/republishing this material for advertising or promotional purposes, creating new collective works, for resale or redistribution to servers or lists, or reuse of any copyrighted component of this work in other works.
DOI: 10.1109/LRA.2026.3752000.}%
\vspace{-1.5em}
}

\markboth{IEEE Robotics and Automation Letters. Preprint Version. Accepted October, 2026}{Biskup \MakeLowercase{\textit{et al.}}: GRF-Aware MPC for Underactuated Upright Physical Assistance}
\maketitle

\begin{abstract}
Many rehabilitation robots execute prescribed motions using reactive tracking controllers that do not explicitly incorporate the user's instantaneous human--ground interaction. This paper presents Ground Reaction Force-Aware Model Predictive Control (GRF-MPC), a framework for upright physical assistance. By integrating live GRF and center of pressure (CoP) measurements directly into a human centroidal dynamics model, GRF-MPC formulates a real-time quadratic program (QP) that optimizes assistive forces and moments over a receding horizon. We deployed GRF-MPC on the Robotic Upright Stand Trainer (RobUST) in an underactuated 4-cable configuration and evaluated it against a fully actuated 8-cable Cartesian impedance baseline in nearly 300 trials across six healthy participants. Across tasks spanning impulsive perturbations, lunge trajectories, and voluntary limits of stability, GRF-MPC reduced peak planar center-of-mass (CoM) error by 19.6\% during perturbations, lowered trajectory tracking error by 24.1\%, required 28.7\% less net mechanical work, and permitted a 45.9\% greater normalized voluntary excursion area. The underlying QP solves in under 1~ms, comfortably enabling 100~Hz replanning and demonstrating that incorporating live human--ground interaction into predictive control yields stronger, more transparent physical assistance with half the actuators.
\end{abstract}


\section{Introduction}
\IEEEPARstart{I}{ndividuals} with mobility impairments resulting from spinal cord injury (SCI), cerebral palsy, and stroke are often wheelchair-bound or need significant assistance to stand. These conditions affect millions worldwide, with approximately 17,000 new SCI cases annually in the United States alone \cite{Ding_Hu_Wang_Kang_Peng_Dong_Li_2022}. Prolonged sitting leads to further muscle deterioration, creating a vicious cycle where weakened muscles make it increasingly difficult to stand leading to further degeneration. To break this cycle, robot-assisted rehabilitation offers customized targeted interventions that enable precise motion control, quantitative assessment capabilities, and reduced physical burden on therapists \cite{healthcare12171720}.

However, many robots for human assistance or rehabilitation, including frame-mounted mechanisms or exosuits, primarily employ control strategies focused on kinematic outcomes or provide joint-level torque assistance. During healthy overground locomotion, humans \rev{generate} stability and propulsion \rev{through ground-contact forces} at the foot-ground interface. When \rev{an assistive} controller injects forces based on trajectory deviations without accounting for these existing ground reaction forces (GRFs), the \rev{resulting net force can produce} center of mass (CoM) accelerations \rev{that are} misaligned with the intended movement direction. \rev{Such mismatch can lead to corrective assistance that opposes natural motion or requires compensatory action from the user or therapist.}

\begin{figure*}[t]
\centering
\includegraphics[width=\textwidth]{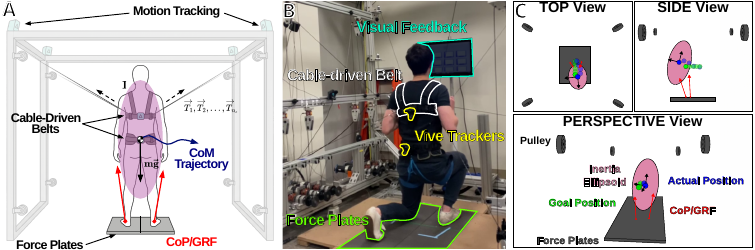}
\caption{(A) Schematic of the Robotic Upright Stand Trainer (RobUST) system highlighting motion tracking, cable actuation, and CoP/GRF measurement. (B) Photograph of the experimental setup showing the user equipped with Vive trackers, standing on force plates, and receiving visual feedback. (C) Real-time graphical interface displaying the user's inertia ellipsoid, target and predicted positions, and force vectors.}
\label{fig:robust}
\end{figure*}

To address the lack of GRF consideration in real-time rehabilitative control, this paper adapts predictive techniques from legged locomotion to physical human-robot interaction. We propose a novel GRF-aware Model Predictive Control (GRF-MPC) framework that proactively optimizes assistance at every time step using real-time centroidal dynamics. \rev{Measured net GRF and center of pressure (CoP) are incorporated directly into the human centroidal dynamics, providing the external human-ground interaction needed to predict CoM evolution without estimating internal joint torques or forecasting a detailed multibody motion. Embedding this measurement-driven model in a constrained quadratic program (QP) inherently handles underactuation while explicitly enforcing cable tension limits.} We experimentally validate this formulation on the Robotic Upright Stand Trainer (RobUST), a custom frame-mounted, cable-driven system (Fig.~\ref{fig:robust}) that applies configurable forces to the user's body with minimal added inertia \cite{8788571,9486923,Santamaria2021}. Compared to an impedance control baseline, GRF-MPC demonstrates improved dynamic tracking and greater efficiency, despite allowing a larger permissible volitional workspace. The principal contributions are:

\rev{
\begin{enumerate}
    \item The formulation of GRF-MPC, a centroidal dynamics controller for human physical assistance in which live GRF measurements serve as the enabling observation that completes the predictive model.
    \item A convex QP formulation with sub-millisecond solve times and 100~Hz replanning that natively accommodates underactuation and tension constraints.
    \item Human-subject validation on the RobUST platform across disturbance rejection, prescribed CoM tracking, net mechanical work, and voluntary workspace, comparing the underactuated 4-cable GRF-MPC against an 8-cable impedance-control baseline.
\end{enumerate}}

\section{Background and Motivation}
\subsection{Rehabilitation Robotics}
Recent systematic reviews emphasize that effective robot-assisted gait rehabilitation relies heavily on intelligent control strategies that encompass assistive, resistive, and perturbative interventions \cite{10517747}. From a clinical perspective, interventions that successfully combine trajectory-tracking with compliant or adaptive assistance demonstrate the highest efficacy for motor function recovery \cite{De_Miguel-Fernandez2023-id}.

To deliver these clinically desired outcomes, many lower-limb rehabilitation or assistive robots predominantly employ a hierarchical control architecture. Systems ranging from treadmill-bound trainers \cite{Baronchelli_Zucchella_Serrao_Intiso_Bartolo_2021} to more recent exoskeletons like WalkON \cite{KIM2025103389} and Atalante \cite{8460647,brunet:hal-04061939} typically operate by first synthesizing offline reference kinematic trajectories known to satisfy Zero-Moment Point (ZMP) stability criterion under quasi-static assumptions.

During online operation, these controllers predominantly adopt a trajectory-tracking strategy that rigidly attenuates state deviations to enforce the reference motion \cite{10242365,7281187}. Because they act to minimize kinematic errors rather than regulating contact forces, they lack explicit, real-time modulation of the underlying ground reaction forces. Consequently, all natural gait variations are treated as tracking errors, resulting in corrective forces that fight human intent.

\subsection{Advances in Legged Robot Control}
While rehabilitation \rev{robotics still relies heavily on prescribed} trajectory tracking, legged robotics has \rev{increasingly adopted} dynamics-aware predictive control \cite{Kuindersma2016}. \rev{A key enabler of highly dynamic behaviors---including} running, bounding, and acrobatic \rev{maneuvers---has been the explicit incorporation} of ground reaction forces into the control formulation.

This transition was largely enabled by utilizing a centroidal dynamics abstraction, which establishes a direct mapping between contact GRFs and the robot's spatial CoM \cite{Orin2013}. By formulating this relationship as an MPC problem, controllers can proactively optimize GRF profiles over a finite time horizon. Di Carlo et al. \cite{8594448}, for example, demonstrated that a convex approximation could be solved at high frequency on the MIT Cheetah 3 to support impressive real-time disturbance rejection and blind terrain traversal.

\subsection{The Rehabilitation Gap}

\rev{Despite these parallel breakthroughs, contemporary reviews of lower-limb exoskeleton and rehabilitation control strategies confirm that the current state-of-the-art heavily relies on impedance, adaptive robust, or assist-as-needed architectures \cite{MDPI_Review_2026, Frontiers_2026}---techniques that remain fundamentally reactive.}

\rev{This lack of predictive foresight is largely due to a key distinction: in legged robotics, GRFs are \textit{control outputs}---the robot generates them through known actuators at sensed contact points. For a human subject, GRFs are \textit{exogenous inputs} generated from the user's volitional muscle activity and can be neither commanded nor predicted by the assistive robot. Examining the centroidal dynamics (Sec.~III-A), the CoM linear and angular acceleration depends on gravity, the robot's exerted forces onto the user, and the user's own GRF. Without measuring GRF, the net force on the CoM is unknown and no predictive model of the human's spatial dynamics can be formulated. GRF measurement thus transforms an otherwise unobservable system into one amenable to predictive control. This is fundamentally distinct from approaches that add force feedback to an existing controller; the live measured GRF is what makes a meaningful predictive model possible.}

By closing the control loop around live GRF \rev{measurements}, GRF-MPC bridges this divide thus advancing \rev{upright physical} assistance from reactive kinematic tracking to proactive, dynamics-aware collaboration.

\section{Methods}

\subsection{Centroidal Dynamics Model}
Regardless of the internal degrees of freedom of any collection of rigid bodies, the dynamics of its CoM and angular momentum, also referred to as centroidal dynamics, are governed entirely by the external forces and the system's inertial properties.

Applying this principle to a human subject of mass $m$ assisted by RobUST, depicted in Fig.~\ref{fig:robust}(A-B), where the external forces consist of gravity $m\mathbf{g}$, the net ground reaction force $\mathbf{f}_{grf} \in \mathbb{R}^3$, and the tension force $\mathbf{f}_{c,i} \in \mathbb{R}^3$ applied by the $i$-th cable at body attachment position $\mathbf{p}_{c,i}$ ($i=1,\dots,n_c$), the Newton-Euler equations of motion are defined as:
\begin{align}
m\ddot{\mathbf{p}} &= m\mathbf{g} + \mathbf{f}_{grf} + \sum_{i=1}^{n_{c}} \mathbf{f}_{c,i}\,, \\
\dot{\mathbf{L}} &= (\mathbf{p}_{cop} - \mathbf{p}) \times \mathbf{f}_{grf} + \sum_{i=1}^{n_{c}} (\mathbf{p}_{c,i} - \mathbf{p}) \times \mathbf{f}_{c,i}\,,
\end{align}
where the vectors $\mathbf{p}, \mathbf{p}_{cop}, \mathbf{p}_{c,i} \in \mathbb{R}^3$ denote the spatial positions of the CoM, the net CoP at the feet, and the $i$-th cable attachment point, respectively. Equation (1) defines the linear dynamics yielding the CoM acceleration $\ddot{\mathbf{p}} \in \mathbb{R}^3$. Equation (2) dictates the rotational dynamics, mapping the net moments about the CoM to the rate of change of centroidal angular momentum $\dot{\mathbf{L}} \in \mathbb{R}^3$.
Unlike the mass $m$, which remains constant, the inertia tensor $\mathbf{I}(t) \in \mathbb{R}^{3 \times 3}$ is time-varying due to continuous pose changes. The exact rotational dynamics are thus:
\begin{equation}
\dot{\mathbf{L}} = \frac{d}{dt}(\mathbf{I}(t)\boldsymbol{\omega}) = \mathbf{I}(t)\dot{\boldsymbol{\omega}} + \dot{\mathbf{I}}(t)\boldsymbol{\omega}\,,
\end{equation}
where $\boldsymbol{\omega}$ is the centroidal angular velocity. In robotics, it is common to know the configuration-dependent inertia $\mathbf{I}(t)$ from design. In human biomechanics, accurately determining this quantity requires estimating inertial parameters for individual body segments and aggregating them into a multibody model \cite{Dempster1955, Zatsiorsky1983}. While body segment inertial parameter estimation is a rigorously established methodology capable of yielding high-fidelity inertia tensors, this approach would still necessitate tracking of multiple body segments, which would add complexity and wearable sensor burden to the human motion assistance system.

In this paper, we approximate the centroidal dynamics by neglecting the time rate of change of the inertia tensor, $\dot{\mathbf{I}}(t)$. Substituting the time-varying $\mathbf{I}(t)$ with a constant, representative inertia tensor $\hat{\mathbf{I}}$ reduces the human model to a Single Rigid Body (SRB) formulation. 

Crucially, the linear dynamics of the CoM (1) are fully captured---the relationship between net external forces and CoM linear acceleration remains exact regardless of subject pose. Because the whole-body CoM is located near the second sacral vertebra (S2) during standing and sacral kinematics exhibit strong correlations to full multi-segment CoM trajectories \cite{Whittle2007, JEONG2018333}, we approximate the CoM position $\mathbf{p}$ using the measured position of the sacrum. \rev{This approximation is appropriate for the upright motions considered here, but is not intended as a general CoM estimator for arbitrary postures.}

To approximate $\hat{\mathbf{I}}$, we model the user as a uniform cylinder defined by mass $m$, height, and a radius equal to half their shoulder width. This rigid-body approximation is physically grounded by typical human mass distribution, where the head and trunk account for 60--70\% of total body mass \cite{DELEVA19961223, Plagenhoef1983}. 

\rev{Under the low angular velocities typical of rehabilitative tasks, the magnitude of the quadratic gyroscopic term $\boldsymbol{\omega} \times (\hat{\mathbf{I}}\boldsymbol{\omega})$ is assumed sufficiently small to omit during linearization.} Together, these structural and kinematic approximations reduce the exact rotational dynamics (3) to the simplified linear form $\dot{\mathbf{L}} \approx \hat{\mathbf{I}}\dot{\boldsymbol{\omega}}$. \rev{To mitigate sensitivity to these approximations during larger limb excursions, orientation cost weights are de-emphasized in the MPC formulation (Sec.~IV-B2).}

\subsection{Convex MPC Formulation}

For cable-driven systems like RobUST, maintaining positive tension ($\mathbf{u} \ge \mathbf{T}_{min}$) is a hard engineering constraint required to prevent slack and ensure safety. Riccati-based methods such as the iterative Linear Quadratic Regulator are highly computationally efficient, but they rely on soft penalties that cannot guarantee strict constraint satisfaction \cite{Howell2019, Tassa2014}. Formulating the finite-horizon control problem as a single convex quadratic program (QP) allows tension limits to be enforced explicitly within the solver.

\subsubsection*{State and Input Definitions}
The system state $\mathbf{x} \in \mathbb{R}^{12}$ comprises the CoM position $\mathbf{p} \in \mathbb{R}^3$ and velocity $\dot{\mathbf{p}} \in \mathbb{R}^3$ expressed in the world frame, alongside the orientation $\boldsymbol{\Theta} \in \mathbb{R}^3$ (parameterized via roll-pitch-yaw Euler angles) and angular velocity $\boldsymbol{\omega} \in \mathbb{R}^3$. The control input $\mathbf{u} \in \mathbb{R}^{n_c}$ consists of the individual tension forces $T_i$ applied by each of the $n_c$ cables. Formally, these vectors are concatenated as:
\begin{equation}
\mathbf{x} = \begin{bmatrix} \mathbf{p}^\top & \boldsymbol{\Theta}^\top & \dot{\mathbf{p}}^\top & \boldsymbol{\omega}^\top \end{bmatrix}^\top, 
\mathbf{u} = \begin{bmatrix} T_1 & T_2 & \dots & T_{n_c} \end{bmatrix}^\top .
\end{equation}

\subsubsection*{Linearization of Continuous Dynamics}

The full nonlinear dynamics of the single rigid body model, incorporating measured GRF, are expressed in control-affine form:
\begin{equation}
\dot{\mathbf{x}} = \mathbf{f}(\mathbf{x}) + \mathbf{B}(\mathbf{x})\mathbf{u}\,,
\end{equation}
where the nonlinear dynamics vector $\mathbf{f}(\mathbf{x})$ and control matrix $\mathbf{B}(\mathbf{x})$ are defined as:
\begin{equation}
\mathbf{f}(\mathbf{x}) = \begin{bmatrix}
\dot{\mathbf{p}} \\
\mathbf{E}(\boldsymbol{\Theta})\boldsymbol{\omega} \\
\mathbf{g} + \frac{1}{m}\mathbf{f}_{grf} \\
\hat{\mathbf{I}}^{-1}\left( -\boldsymbol{\omega} \times \hat{\mathbf{I}}\boldsymbol{\omega} + (\mathbf{p}_{cop} - \mathbf{p}) \times \mathbf{f}_{grf} \right)
\end{bmatrix} \,,
\end{equation}
\rev{\begin{equation}
\mathbf{B}(\mathbf{x}) = \begin{bmatrix}
\mathbf{0}_{3 \times n_c} \\
\mathbf{0}_{3 \times n_c} \\
\dots \frac{1}{m}\mathbf{n}_i(\mathbf{x}) \dots \\
\dots \hat{\mathbf{I}}^{-1} \left((\mathbf{p}_{c,i}(\mathbf{x})-\mathbf{p}) \times \mathbf{n}_i(\mathbf{x})\right) \dots
\end{bmatrix}_{i=1 \dots n_c} \,.
\end{equation}}

In (6), $\mathbf{E}(\boldsymbol{\Theta})$ maps body angular rates to Euler angle rates, and $\mathbf{n}_i, \mathbf{p}_{c,i}(\mathbf{x})-\mathbf{p}$ from (7) represent the $i$-th cable's unit vector (pointing away from the belt) and moment arm relative to the CoM.

Because future user GRF and CoP cannot be reliably predicted, treating these external forces as constant over the prediction horizon naturally aids in reducing the system's nonlinearities to an affine form. With the external forces fixed, we linearize the continuous dynamics by evaluating the kinematic mapping $\mathbf{E}(\boldsymbol{\Theta})$ and control matrix $\mathbf{B}(\mathbf{x})$ at the current measured state $\mathbf{x}_0$. By omitting the quadratic gyroscopic term $\boldsymbol{\omega} \times (\hat{\mathbf{I}}\boldsymbol{\omega})$ (as physically justified in Sec. III-A), we avoid iterative or sequential nonlinear programming schemes and obtain the following affine system, reducing the trajectory optimization to a single QP: 
\begin{equation}
\dot{\mathbf{x}} = \underbrace{\begin{bmatrix}
\mathbf{0}_{3\times3} & \mathbf{0}_{3\times3} & \mathbb{I}_{3\times3} & \mathbf{0}_{3\times3} \\
\mathbf{0}_{3\times3} & \mathbf{0}_{3\times3} & \mathbf{0}_{3\times3} & \mathbf{E}(\boldsymbol{\Theta}_{0}) \\
\mathbf{0}_{3\times3} & \mathbf{0}_{3\times3} & \mathbf{0}_{3\times3} & \mathbf{0}_{3\times3} \\
\mathbf{0}_{3\times3} & \mathbf{0}_{3\times3} & \mathbf{0}_{3\times3} & \mathbf{0}_{3\times3}
\end{bmatrix}}_{\mathbf{A}} \mathbf{x} + \mathbf{B}(\mathbf{x}_0) \mathbf{u} + \mathbf{d} \,.
\end{equation}
The drift vector $\mathbf{d}$ groups the external forces---gravity and GRF---into a single affine term:
\begin{equation}
\mathbf{d} = \begin{bmatrix} \mathbf{0}_{6\times1} \\ \mathbf{g} + \frac{1}{m}\mathbf{f}_{grf} \\ \hat{\mathbf{I}}_0^{-1}((\mathbf{p}_{cop} - \mathbf{p}_0) \times \mathbf{f}_{grf}) \end{bmatrix}\,.
\end{equation}

\subsubsection*{Discrete Time Dynamics}
To predict system evolution for digital implementation over a horizon $N$ with timestep $\Delta t$, we use a simple Forward Euler discretization. Since both $\mathbf{A}$ and $\mathbf{B}(\mathbf{x})$ are treated as constant over the planning horizon, the discrete dynamics are given by the difference equation:
\begin{equation}
\mathbf{x}_{k+1} = (\mathbb{I}_{12\times12} + \mathbf{A}\Delta t) \mathbf{x}_k + \mathbf{B}(\mathbf{x}_0)\Delta t\mathbf{u}_k + \mathbf{d} \Delta t \,.
\end{equation}

\subsubsection*{QP Construction}
We seek the optimal input sequence $\mathbf{U} \in \mathbb{R}^{n_c N}$ that minimizes the following objective function subject to actuation constraints:
\begin{equation}
\begin{aligned}
J(\mathbf{U}) &= \sum_{k=0}^{N-1} \left( \|\mathbf{x}_{k+1} - \mathbf{x}_{ref,k+1}\|_{\mathbf{Q}}^2 + \|\mathbf{u}_k\|_{\mathbf{R}}^2 \right) \\
\text{s.t.} \quad & \mathbf{T}_{min} \le \mathbf{u}_k \le \mathbf{T}_{max} \,,
\end{aligned}
\end{equation}
where the notation $\|\mathbf{z}\|_\mathbf{W}^2$ denotes the weighted squared Euclidean norm $\mathbf{z}^\top \mathbf{W} \mathbf{z}$, and $\mathbf{Q} \in \mathbb{R}^{12 \times 12}$, $\mathbf{R} \in \mathbb{R}^{n_c \times n_c}$ are diagonal weighting matrices. To solve this efficiently, the objective function (11) is converted into a standard condensed quadratic program by expressing the state trajectory as an affine function of the initial state and input sequence, allowing rapid solution via standard convex solvers.

\subsection{System Architecture}

\begin{figure}[t]
\centering
\includegraphics[width=1\linewidth]{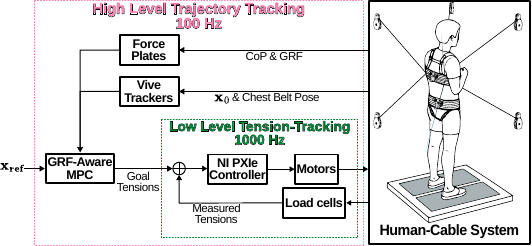}
\caption{Hierarchical control architecture for GRF-MPC on RobUST. The high-level controller implemented in C\# uses tracker and force plate data to solve the convex QP problem in real-time. The resulting optimal tension commands are transmitted asynchronously to the low-level NI PXIe system, which tracks the tension via load cell feedback.}
\label{fig:system_architecture}
\end{figure}

The control architecture is organized into a hierarchical framework separating high-level trajectory planning and optimization from low-level actuator control (Fig.~\ref{fig:system_architecture}). This separation allows for computationally intensive optimization to occur in real-time, while ensuring safety and tracking performance through low-level actuator loops.

\subsubsection*{Hardware and Sensing}
To capture the requisite kinematic and dynamic state variables, two HTC Vive trackers (90~Hz) provide spatial localization of the chest-mounted belt end-effector and the user's sacrum, while floor-mounted force plates (100~Hz) measure the GRFs and CoP positions (Fig.~\ref{fig:robust}(B)). The use of a physical tracker for end-effector localization ensures reliable positioning even in the presence of cable slack. The belt is modeled as a rigid body with known cable attachment points in the end-effector frame.

\subsubsection*{High-Level Control}
The high-level controller is implemented in C\# to leverage native compatibility with the HTC Vive interface via SteamVR. It serves as the MPC planner, using the last known readings on GRFs and CoM position and velocity to compute the cable tensions needed to manipulate the centroidal dynamics. System states, including the user's inertia ellipsoid and dynamic force vectors, are monitored via a real-time graphical interface (Fig.~\ref{fig:robust}(C)). \rev{At each 100~Hz replan cycle, the planner constructs a 0.5~s look-ahead ($N=10$, $\Delta t = 0.05$~s) and outputs the first element of the optimal tension sequence.} The optimization problem is solved using a generic off-the-shelf Interior Point Method (IPM) convex QP (ALGLIB \texttt{minqp} subpackage). 

\subsubsection*{Low-Level Control}
The optimal tension commands computed by the high-level planner are transmitted over an asynchronous TCP/IP interface to a National Instruments PXIe-8880 controller. This low-level controller operates in hard real-time, executing a tuned PID control law at $1$~kHz to track the commanded tension setpoints using live load cell feedback. \rev{Treating the high-bandwidth PID loop as an ideal tension actuator is a standard paradigm in agile robotics \cite{Kamel2017, Schwarting2018}, allowing the low-level loop to absorb transient forces (e.g., cable slack recovery) at timescales faster than the MPC horizon.}

\section{Experimental Validation}

To evaluate the clinical potential and dynamic performance of GRF-MPC, a comparative study was conducted on the RobUST platform. The primary objective was to demonstrate whether a predictive, underactuated 4-cable MPC framework can outperform a fully-specified 8-cable reactive impedance controller---representing the current standard in robotic rehabilitation---by providing superior assistance, better transparency, and requiring less net control effort.

\rev{This comparison is intentionally asymmetric. The impedance baseline requires 8 cables to guarantee force-closure, whereas GRF-MPC natively accommodates underactuation. This contextualizes the results of Tasks 1 and 2: despite operating with fewer actuators and inherently less force authority, GRF-MPC achieves superior disturbance rejection and tracking precision.}

\subsection{Experimental Setup}
The study recruited a total of 6 healthy individuals (three male and three female; age: 26.66 $\pm$ 1.97 years, height: 167 $\pm$ 7.8 cm). All participants provided written informed consent under a protocol approved by the Columbia University Institutional Review Board (AAAR6780). The system was tested under three distinct control conditions: (1) \textit{Transparent Mode}, where the robot applies only a minimal 10~N constant tension per cable to prevent slack, serving as a minimally assisted baseline approximating zero-impedance; (2) \textit{8-Cable Impedance Control}, representing the fully-actuated state-of-the-art approach; and (3) \textit{4-Cable GRF-MPC}, utilizing the proposed predictive framework.

\subsection{Controller Formulations and Tuning}

\subsubsection{Baseline: 8-Cable Impedance Control}
The CoM reference state is continuously mapped to the belt frame, and a standard Cartesian impedance law generates a restoring \rev{wrench (combined force-torque vector)} $\mathbf{W}_{des}$ proportional to this kinematic deviation. Base stiffness parameters (500~N/m horizontally, 1100~N/m vertically) were established using a 58~kg pilot subject and scaled linearly by participant mass, with minimal damping to balance strict tracking against user comfort. Finally, $\mathbf{W}_{des}$ is allocated to individual cable tensions via a QP that minimizes squared tensions while strictly enforcing wrench execution within physical tension limits \rev{($T_{min} = 10$~N, $T_{max} = 200$~N)} (Fig.~\ref{fig:3}(A), left).

\subsubsection{Proposed: 4-Cable GRF-MPC}
\begin{figure}[t]
\centering
\includegraphics[width=1\linewidth]{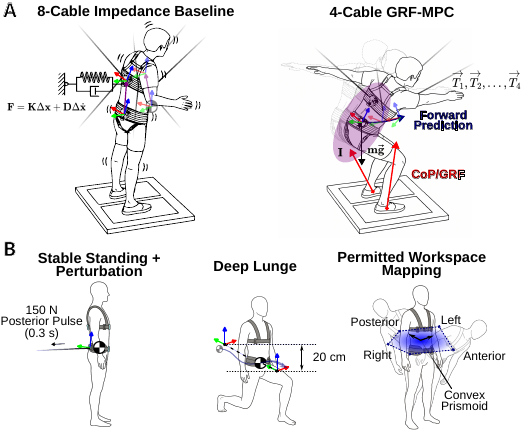}
\caption{(A) Illustration of the fully actuated impedance control baseline, contrasted with the underactuated 4-cable GRF-MPC. (B) Illustration of experimental tasks: Stable Standing under perturbation (left); Deep Lunge tracking (middle); and Permitted Workspace Mapping via voluntary chest displacement (right).}
\label{fig:3}
\end{figure}

GRF-MPC was deployed in a 4-cable configuration (Fig.~\ref{fig:3}(A), right). \rev{Tension constraints were strictly enforced within the QP solver. The MPC weights were tuned iteratively via the real-time graphical interface (Fig.~\ref{fig:robust}(C)) and are summarized in Table~\ref{tab:params}. To balance tracking accuracy with natural kinematics, the state cost matrix $\mathbf{Q}$ heavily penalized horizontal trajectory deviations while relaxing the vertical constraint to permit natural CoM oscillations. Velocity damping was scaled an order of magnitude lower, and orientation states were deliberately de-emphasized to mitigate SRB model mismatch. The input cost was set to a minor regularization value. Notably, the exact same MPC weights were used for all six subjects without any mass scaling.}

\rev{\begin{table}[t]
\centering
\caption{Controller Parameters}
\label{tab:params}
\footnotesize
\begin{tabular}{l l}
\hline
\textbf{Parameter} & \textbf{Value} \\
\hline
\multicolumn{2}{l}{\textit{4-Cable GRF-MPC}} \\
$N$ (prediction horizon) & 10 steps \\
$\Delta t$ (discretization) & 0.05 s \\
$Q_{x,y}$ & 500 \\
$Q_z$ & 100 \\
$Q_{\dot{p}}$ & 50 \\
$Q_{\Theta,\omega}$ & 1 \\
$R$ & $10^{-3} \mathbb{I}$ \\
$T_{min}$, $T_{max}$ & 10, 200 N \\
Replan rate & 100 Hz \\
\hline
\multicolumn{2}{l}{\textit{8-Cable Impedance Control}} \\
$K_{horiz}$ (base, 58 kg pilot) & 500 N/m \\
$K_{vert}$ (base, 58 kg pilot) & 1100 N/m \\
Mass scaling & $K_i = K_{base} \cdot m / 58$ \\
$T_{min}$, $T_{max}$ & 10, 200 N \\
\hline
\end{tabular}
\end{table}}

\subsection{Task Protocols and Reference Generation}
Subjects performed three specific tasks designed to test disturbance rejection, dynamic tracking, and workspace transparency as illustrated in Fig.~\ref{fig:3}(B). For all tasks, the reference state $\mathbf{x}_{ref}$ and reference input $\mathbf{u}_{ref}$ were updated at the 100~Hz planning frequency. \rev{Subjects performed all three tasks sequentially under each control condition (Transparent, 4-Cable MPC, 8-Cable Impedance) with brief 5--10~s rest periods between condition switches.}

\begin{enumerate}
    \item \textbf{Stable Standing + Perturbation (10 trials):} The subject maintained a neutral standing posture. The reference state $\mathbf{x}_{ref}$ was defined as the static initial CoM position. An unanticipated 150~N posterior pulse over 0.3~s was applied via a cable attached to the pelvic belt, testing the controller's ability to arrest momentum. \rev{Perturbations were delivered at roughly 10-second intervals.}
    \item \textbf{Deep Lunge (5 trials):} The subject performed a controlled lunge. The reference trajectory $\mathbf{x}_{ref}$ was linearly interpolated from the neutral standing height to a geometric waypoint requiring a 20~cm vertical drop, and back. Rather than providing a biologically optimal path, this dynamically infeasible trajectory exposes systematic differences in handling aggressive tracking demands. \rev{This represented the largest standardized dynamic movement achievable within the force plate area.}
    \item \textbf{Permitted Workspace Mapping (10 cycles):} The subject began in a neutral posture and voluntarily leaned their torso as far as possible in anterior, posterior, and lateral directions (a ``star'' pattern). With $\mathbf{x}_{ref}$ remaining at neutral and the chest tracker driving displacement, this mapped the maximum workspace permitted before restorative forces compelled a return.
\end{enumerate}

\rev{\subsection{Outcome Measures and Statistical Analysis}}

\rev{For Task 1 (perturbation), transient balance recovery was quantified by two metrics: (1) peak error, defined as the absolute difference between the maximum planar CoM excursion and the steady-state position (averaged over a 4--8~s window post-peak), and (2) settling time, defined as the duration required for the CoM to enter and remain within a 5\% error band of this steady-state position.}

\rev{For Task 2 (lunge), planar tracking accuracy was measured as the CoM planar error relative to the reference trajectory. Robotic effort was quantified as cumulative mechanical work: at each timestep, instantaneous power delivered to the user was computed as $P = \mathbf{F}_{cmd} \cdot \mathbf{v}_{belt} + \boldsymbol{\tau}_{cmd} \cdot \boldsymbol{\omega}_{belt}$, and integrated over the task duration to quantify the total work imparted onto the user, independent of cable count or controller type.}

\rev{For Task 3 (workspace), the permissible excursion area was defined as the convex prismoid enclosing the chest tracker trajectory during maximal voluntary leanings, normalized by the square of the subject's height.}

\rev{Statistical significance was evaluated using a Linear Mixed-Effects (LME) model given the high inherent variance of human biomechanical data. LME models are increasingly utilized in biomechanical and kinesiological research because they account for individual baseline differences by modeling random subject intercepts and avoid the strict sphericity assumptions required by traditional repeated-measures ANOVA \cite{Mota2023-gj, ijerph18020392}. The controller condition (Transparent, Impedance, MPC) was modeled as a fixed effect to isolate the algorithmic performance. To account for differing baseline physical and neuromuscular profiles, subject was modeled as a random effect (varied intercepts). Results are reported with coefficient estimates ($\beta$) indicating the change relative to the specified baseline, evaluated at a significance level of $p = 0.05$.}

\section{Results}

\rev{\begin{figure}[tp!]
    \centering
    \includegraphics[width=\linewidth]{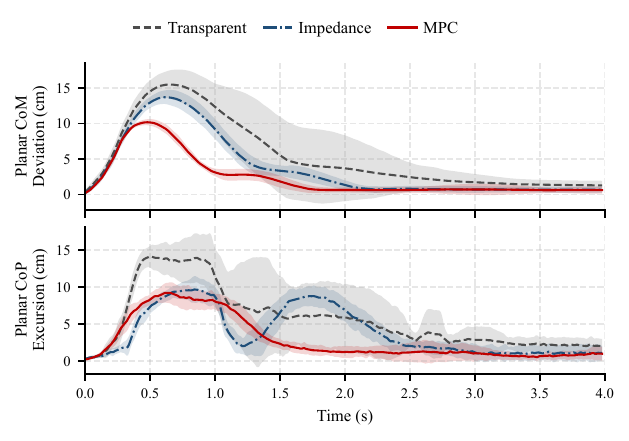}
    \caption{Representative perturbation response for a single subject. (Top) Planar CoM position vs.\ time under Transparent, Impedance, and MPC conditions. (Bottom) Planar CoP excursion measured by the force plates during the same trial.}
    \label{fig:timeseries}
\end{figure}}

\begin{figure*}[tp!]
    \centering
    \includegraphics[width=\textwidth]{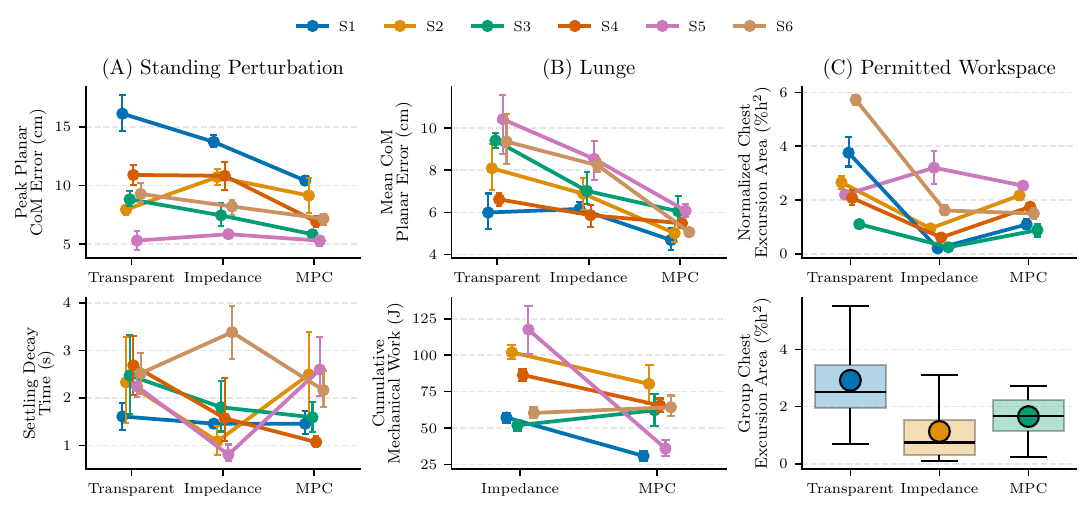}
    \caption{Experimental results across all three tasks. Error bars indicate Standard Deviation. (A) Task 1: Standing Perturbation — comparison of peak planar CoM error (top) and settling decay time (bottom) for subjects S1--S6 under Transparent, Impedance, and MPC control modes. (B) Task 2: Lunge tracking efficiency — CoM planar error (top) and cumulative mechanical work (bottom). (C) Task 3: Permitted workspace mapping — normalized excursion area demonstrating the MPC's ability to preserve a significantly larger functional workspace compared to rigid impedance control.}
    \label{fig:fig_results}
    \label{fig:fig_standing_plots}
    \label{fig:fig_lunge_plots}
    \label{fig:fig_excursion_plots}
\end{figure*}

\subsection{Task 1: Disturbance Rejection}
\rev{A representative time-domain perturbation response is shown in Fig.~\ref{fig:timeseries}.} Group-level transient balance recovery performance is presented in Fig.~\ref{fig:fig_standing_plots}(A).

The Transparent baseline yielded a mean peak error of 0.097 m and a settling time of 1.27 s. The 8-cable Impedance controller reduced the settling time by 38.6\% ($\beta = -0.502$, $p < 0.001$), but did not yield a statistically significant reduction in peak error ($p = 0.814$). The 4-cable GRF-MPC significantly reduced both peak error by 19.6\% ($\beta = -0.019$, $p < 0.001$) and settling time by 48.0\% ($\beta = -0.619$, $p < 0.001$) relative to the baseline.

When the active controllers were compared directly, the MPC achieved a 19.6\% reduction in peak error over the Impedance controller ($\beta = -0.018$, $p < 0.001$), while yielding a comparable settling time ($p = 0.245$).

\subsection{Task 2: Dynamic Tracking}
Planar tracking accuracy and robotic effort during the 20 cm CoM drop are presented in Fig.~\ref{fig:fig_lunge_plots}(B). Relative to the Transparent baseline tracking error of 8.32 cm, the Impedance controller provided a 14.4\% reduction ($\beta = -1.20$ cm, $p < 0.001$). GRF-MPC yielded a 35.1\% reduction relative to the baseline ($\beta = -2.92$ cm, $p < 0.001$), corresponding to a 24.1\% improvement in tracking precision over the Impedance controller ($\beta = -1.72$ cm, $p < 0.001$). 

Statistical comparison of \rev{cumulative mechanical work} confirmed that the MPC commanded 28.7\% less total mechanical work than the Impedance controller ($\beta = -22.77$ J, $p < 0.001$).

\subsection{Task 3: Permitted Workspace Mapping}

The Transparent condition permitted the largest kinematic workspace (Fig.~\ref{fig:fig_excursion_plots}(C)), averaging a height-normalized area of $2.92$ \%(height$^2$). Both controllers significantly restricted this area ($\beta = -1.785$, $p < 0.001$ and $\beta = -1.265$, $p < 0.001$), but GRF-MPC permitted 45.9\% more normalized excursion area than the Impedance controller ($\beta = 0.520$, $p = 0.002$).

\section{Discussion}

The experimental data demonstrates a systematic superiority of GRF-MPC over the state-of-the-art baseline, despite underactuation. The transient perturbation data (Task 1) exposes a critical biomechanical limitation of many existing exoskeleton and rehabilitation robots. While it succeeds in quickly returning the user to a setpoint, it fails to meaningfully mitigate the initial peak excursion. In a purely reactive architecture, restorative forces are strictly proportional to state error; the user remains unassisted during the initial impact dynamics. Mitigating this initial displacement would require increasing stiffness gains, which inherently compromises dynamic comfort and compliance. GRF-MPC achieves a nearly 20\% reduction in peak displacement compared to the impedance baseline. This reduction is achieved alongside lower energy metrics and higher compliance in subsequent tasks, demonstrating algorithmic superiority rather than mismatched or weak tuning.

Lunging motion (Task 2) further confirms this performance advantage during voluntary motion. The lunge task establishes superior tracking performance by GRF-MPC. By actively predicting the required trajectory forces, GRF-MPC achieved 24.1\% better tracking precision than the 8-cable impedance controller. Crucially, the concurrent 28.7\% reduction in commanded mechanical work proves this precision is not the result of an under-tuned impedance baseline. The MPC intelligently coordinates the underactuated tensions to track the reference more efficiently, achieving tighter adherence while expending less robotic effort.

Finally, the Permitted Workspace Mapping (Task 3) contextualizes the stiffness-compliance trade-off. Both active controllers constrain the operational envelope relative to the unassisted Transparent condition. However, GRF-MPC---which already established superior tracking and disturbance rejection---gracefully manages underactuation to expand the dynamically stable workspace, permitting 45.9\% more volitional excursion area than the impedance condition. This was achieved while executed in a 4-cable configuration where standard pseudo-inverse Cartesian control is mathematically infeasible.

\rev{Several limitations of the current implementation warrant consideration and motivate future improvements. First, the SRB approximation relies on a coarse estimated inertia matrix $\hat{\mathbf{I}}$ and neglects configuration-dependent rotational dynamics. Second, treating the measured GRFs as constant over the prediction horizon enables a convex QP formulation, but compromises prediction accuracy when contact forces vary substantially. Extending the horizon indefinitely is therefore not necessarily advantageous, since longer predictions extrapolate stale contact-force measurements. Future iterations could address these linearizations by quantifying bounded deviations of $\hat{\mathbf{I}}(t)$ and $\mathbf{f}_{grf}(t)$ to implement Robust or Tube MPC techniques \cite{Pandala2022}.

Third, the present validation was limited to healthy young adults; performance and safety in populations with neurological or musculoskeletal impairments remain to be established. Finally, the system currently relies on floor-mounted force plates to measure GRF and CoP directly. Because the control formulation requires measurement or estimates of human--ground interaction, deployment outside instrumented environments will require alternative wearable sensing paradigms or real-time GRF estimation techniques.}

Additionally, to provide an equivalent comparison to trajectory tracking in exoskeletons and rehabilitation robots, the current controller forces a time-based progression along $\mathbf{x}_{ref}$. However, clinical literature indicates that allowing rehabilitation patients to self-pace their progression through a path in space increases active participation and motor learning \cite{Duschau-Wicke2010-bs, Marchal-Crespo2009-vv}. A promising future extension is the adaptation of Model Predictive \textit{Contouring} Control (MPCC) \cite{Liniger2015}. MPCC decouples path geometry from time by penalizing contour deviations while treating path progress as an optimization variable. By relaxing or removing the progress-maximizing term within the MPCC objective, the controller could provide dynamically safe, spatial-only boundaries, enabling fully patient-driven pacing during rehabilitation tasks.

\section{Conclusion}
Current robotic human motion assistance paradigms fundamentally rely on reactive kinematic controllers that neglect the dynamic influence of immediate ground reaction forces. To address this gap, this paper introduced GRF-MPC on the RobUST platform, translating centroidal predictive control methods from legged robotics to upright physical assistance. 

Experimental validation confirmed that explicitly modeling the relationship between GRFs and spatial center of mass dynamics enables proactive physical support. Implemented in an underactuated 4-cable configuration, GRF-MPC systematically outperformed a fully actuated 8-cable Cartesian impedance baseline. GRF-MPC safely arrested transient momentum with lower peak deviations during unexpected perturbations, achieved superior dynamic tracking precision while exerting significantly less mechanical work, and preserved a substantially larger kinematically transparent workspace during voluntary motion. Ultimately, this work demonstrates that shifting from rigid path-enforcement to dynamically consistent, GRF-aware control yields more fluid, efficient, and biomechanically accurate patient support.

\bibliographystyle{IEEEtran}
\bibliography{references}    

\end{document}